# Text Recognition in Scene Image and Video Frame using Color Channel Selection

[a]Ayan Kumar Bhunia, [b]Gautam Kumar, [b]Partha Pratim Roy*, [b]R. Balasubramanian, [c]UmapadaPal

[a]Department of ECE, Institute of Engineering & Management, Kolkata, India
[b]Department of CSE, Indian Institute of Technology Roorkee, India
[c]CVPR Unit, Indian Statistical Institute, Kolkata, India
*email: proy.fcs@iitr.ac.in

**Abstract**

In recent years, recognition of text from natural scene image and video frame has got increased attention among the researchers due to its various complexities and challenges. Because of low resolution, blurring effect, complex background, different fonts, color and variant alignment of text within images and video frames, etc., text recognition in such scenario is difficult. Most of the current approaches usually apply a binarization algorithm to convert them into binary images and next OCR is applied to get the recognition result. In this paper, we present a novel approach based on color channel selection for text recognition from scene images and video frames. In the approach, at first, a color channel is automatically selected and then selected color channel is considered for text recognition. Our text recognition framework is based on Hidden Markov Model (HMM) which uses Pyramidal Histogram of Oriented Gradient features extracted from selected color channel. From each sliding window of a color channel our color-channel selection approach analyzes the image properties from the sliding window and then a multi-label Support Vector Machine (SVM) classifier is applied to select the color channel that will provide the best recognition results in the sliding window. This color channel selection for each sliding window has been found to be more fruitful than considering a single color channel for the whole word image. Five different features have been analyzed for multi-label SVM based color channel selection where wavelet transform based feature outperforms others. Our framework of color channel selection is script-independent. It has been tested in English (Roman) and Devanagari (Indic) scripts. We have tested our approach on English datasets (ICDAR 2003, ICDAR 2013, MSRA-TD500, IIIT5K, SVT, YVT) publicly available for both video and scene images. For Devanagari script, we collected our own dataset. The performances obtained from experimental results are encouraging and show the advantage of the proposed method.[1]

---

## 1. Introduction

In the field of computer vision problems, text detection and recognition have gained plenty of attentions in recent years. The reason for such interest is due to easy availability of large amount of digital information from videos and scene images which contain very useful information like street name, location's address, traffic warning etc. Therefore, text extraction and recognition from this digital information are very effective and important in different text-based application like data mining, retrieval of images/videos from the large database etc. The major problem in recognizing text from natural scene images and video frames are low-resolution image or frames with different fonts and size of characters, images with a complex background, variant alignment and color etc. Because of these factors, existing commercial document OCR systems do not work well for recognition of text from natural scene images or video frames [3, 40, 60].

Text recognition methods can be grouped into three categories where (i) some approaches recognize the text by using segmentation of text and proposing the classifier training with their own features [3], (ii) methods in the second category recognizes the text without segmentation of the text by using a framework which is based on multiple hypotheses [5], and (iii) the methods of the third category enhance the text to increase the recognition rate by using binarization of scene images [6]. Each of these three categories has their own limitations. Approaches of the first category work well only for data from specific scripts because they need training from their own samples and a classifier to recognize the text based on this training. The methods of the second category need multiple hypotheses to set the thresholds but it is not clear how to draw different hypotheses to set specific thresholds. However, methods of the third category do not need any classifier and hypotheses to set some thresholds and it also enhances the recognition rate through binarization. However, the approaches due to third category do not provide satisfactory recognition performance for low-resolution scene images/video frames. Roy et al. [7] proposed a text binarization approach which improved the recognition rate of natural scene text. It concludes that if the binarization of segmented text line from natural scene images can be improved, then the recognition performance can automatically be improved with available OCRs. However, these methods do not work well for curved text and focused only on horizontal scene text. There exist some approaches [8, 9] for multi-oriented texts in graphical documents. For instance, Pal et al. [9] proposed an approach for recognition of the character in multi-orientation and multi-scaled document. However, again these methods are only well suited for scanned images and do not provide good performance in case of natural scene images or video frames.

Chattopadhyay et al. [17] claimed that a single binarization method may not be good for recognition of text from document image. They proposed a robust OCR in document images by selecting appropriate binarization method. For this purpose, a Support Vector Machine (SVM) based approach was used to select one of the binarization techniques suitable for a particular image. Different image properties like mean, variance, skewness and histogram feature from Hue channel were considered as features. However, they considered only single label classification approach though more than one binarization methods might provide correct recognition result. Inspired with this idea, in this paper we explore different color channels to recognize text from scene image/video frames. A similar color selection approach has been adopted in [44] for object classification in Animal with Attribute (AwA) dataset. The authors in [44] proposed a greedy algorithm using Discriminative and Reliable Attribute Learning (DRAL) which selects a subset of attributes to maximize an objective function. Liu et al. [55] proposed a unified framework for attribute prediction which exploits the relation between attributes to boost the performance of attribute-based learning methods. In this study we have used color selection approach to improve text recognition in scene image and video frame. We show how the complex binarization problem can be avoided by selecting a proper color channel.
In the following, we describe in brief the color spaces which are used in our framework.

**Different Color spaces and their channel description:** Color space [52] is a model for visualizing the color which represents colors using different intensity values. There exist a number of color spaces, e.g. RGB, HSV, YCrCv, CMYK, CIELAB, etc. In this paper, we have considered three color spaces, namely RGB, YCbCr and HSV.
The most commonly used color model is RGB which contains three color channels, namely Red (R), Green (G) and Blue (B) channels. The ranges of the intensity values at each channel lies between 0 and 255. In most of the digital image processing module, RGB color space is vastly used. HSV stands for Hue, Saturation, and Value. Hue represents the dominant color which is used to distinguish colors. Information regarding the color is contained in Hue channel. Saturation channel contains the information regarding the amount of white color is added to the pure color. The Value provides a measure of the intensity of a color [52]. RGB components are separated into luminance and chrominance information after converting to $YC_bC_r$ color space. Y is calculated from the weighted sum of RGB values. $C_b$ is the difference between blue and luminance component, and $C_r$ is the difference between red and luminance component [52].

In this paper, we propose a novel approach of text recognition from scene images and video frames using an automatic selection of color channel and then selected color channel is used for recognition. Given a text image, our framework analyses the image properties of different color channels and then a multi-label SVM classifier is applied to select the color channel that will provide the best recognition results. Once the color channel is selected, Hidden Markov Model (HMM) based classifier is used on this channel for scene text recognition where Pyramidal Histogram of Oriented Gradient (PHOG) features from sliding windows are extracted and fed into HMM. From each sliding window of a color channel, our color-channel selection approach analyzes the image properties from the sliding window and then a multi-label Support Vector Machine (SVM) classifier is applied to select the color channel that will provide the best recognition results in the sliding window. In Fig.1 some examples of word images from scene images and video frames are given where it is noted that a word image may get recognized correctly in some color channels whereas it does not work well using other color channels. Using our proposed color channel selection based approach we can select the correct color channel automatically and hence recognition performance can be improved significantly. Instead of considering a single color channel for an image, we have extended the color channel selection at sliding window level where a color channel is selected for each sliding window and PHOG feature is extracted using the selected channel of the window. Our proposed framework of color channel selection is script-independent and we have tested in Latin and Devanagari (Indic) scripts to show the efficiency.

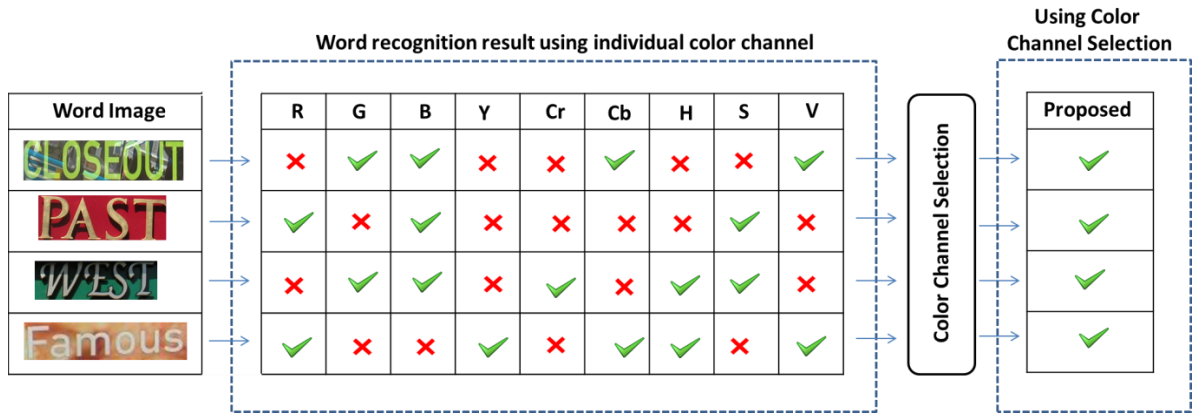

**Fig.1.** Word recognition results using individual color channels and our proposed color channel selection approach. Note that, all these images were not recognized by a single color channel. Our proposed framework selects the color channel automatically from the image and that color channel is used for HMM-based text recognition purpose.

The contributions of the paper are as follows; (1) we propose a color channel selection algorithm using five different image property based features for the purpose of text recognition in scene and video frame. A comparative study has been done to evaluate the performance of each feature. (2) Color channel selection has been extended to individual sliding window level of the text image and further improvement has been found than considering a single color channel. (3) Extensive experiment and analysis have been done in multiple scripts (Latin (English) and Devanagari) to show the improvement of recognition performance using selective color channel rather than using traditional approaches of complex binarization algorithm.

The rest of the paper is organized as follows. In Section 2, we discuss some related works developed for scene/video image text recognition. In Section 3, the proposed framework for color channel selection has been described. Section 4 describes the general HMM based word recognition framework. In Section 5, we provide the experiment setup and discuss performance results in details. Finally, the conclusion is given in Section 6.

## 2. Related Work

Text recognition of natural scene images and video frames is performed in four steps, namely, localization, extraction, binarization and recognition [10]. Several works have been done in the field of text detection [3, 12, 16]. Similarly, many methods exist for text binarization [14, 15]. Recognition of text can be performed either by using segmentation or without using segmentation. Recognition with segmentation has incurred many problems like over segmentation of images or under-segmentation of images and due to which significant information may be lost which leads to poor recognition rate [53]. For instance, sometimes proper segmentation of scene images may not be possible which leads to an improper recognition of the character of scene images like 'm' can be recognized as 'n' or 'w' as 'v'[53]. Wang et al. [15] performed text detection and localization from scene images and video frames in two ways: (i) Region-based detection uses the different properties between background and candidate text region of images to extract the text. (ii) Texture based approach uses different texture feature of text to separate text from the background. These are used to improve text detection approach. Yang et al. [26] proposed a fast localization verification method using edge based multi scale text detector. To eliminate the false alarm they have used SVM and stroke width transform verification procedure and finally uses skeleton based binarization technique for recognition of text. Jain et al. [25] proposed another end to end system for text detection using rich descriptor to detect the positive candidate region with the help of SVM. Then detected region will be binarized, filtered and passed through Hidden Markov Model (HMM) based OCR for text recognition.

**Text recognition from scene images:** A number of works have been proposed for the recognition of text from natural scene images. These methods are mainly based on the concepts of binarization, edge detection and spatial frequency analysis of images. For example, a system [42] was proposed which performs detection, segmentation and recognition of images in a single framework which helps in accommodating contextual relationships. Huang et al. [31] worked on background invariant features to detect the text region from natural scene images. They mainly followed the top-down approach of text recognition. Wang et al. [15] have proposed a method for scene text recognition which uses the traditional visual features like HOG (Histogram of Oriented Gradient). This method enhanced the recognition accuracy using lexicons. The work proposed in [1] uses HMM for multi-oriented scene text recognition. These methods did not exploit color channel to improve the text recognition performance. Gonzalez et al. [16] proposed a text detection and recognition framework on traffic panels. They performed the blue and white color segmentation on the traffic panel images to find the interested key points in the image and then perform the character recognition on interested area. Text recognition from traffic signs framework was also proposed by Greenhalgh et al. [38]. They used MSERs and HSV color thresholding to detect the text region. To reduce the number of false alarms they applied constraints on structural and temporal information on the candidate region before applying recognition method on them. Shivakumara et al. [48] proposed a text detection approach using feature based on Pseudo-Zernike moments, Fourier and Polar descriptor followed by SVM based classification. Then text recognition was performed using OCR after binarization. Jaderberg et al. [49] proposed an end-to-end system for localizing and recognizing texts from natural scene images. For localizing the texts they proposed a system based on region proposal mechanism and used deep convolutional neural network for recognition purpose. They proposed another approach using deep learning for text spotting [50]. Alsharif et al. [43] proposed an approach for word recognition from natural scene images which is a lexicon free segmentation based approach. They used Hybrid HMM/Maxout model for segmentation of words into characters and then constructed cascade for removal of false characters and finally used a variant of the Viterbi algorithm for recognition purpose.

**Text Recognition from video frames:** Text recognition in video frames is more challenging than that of natural scene images. Only a few pieces of work [20, 21, 40] exist for text recognition from video frames. The difficulties in video text analysis are due to low resolution frames, complex background, variation in color, non-uniform illumination, font, font size, camera motion and other blurring artifacts, etc. Hence, currently available OCR gives poor performance in video text images. Most of the recognition methods perform the pre-processing of images before applying the standard segmentation algorithm on video frames. Many pieces of work [20, 21, 28] have been proposed previously on binarization of images. Saidane and Gracia [28] proposed a binarization method based on a convolutional network for color text area of video frames and its performance depends on the amount of training sample used. In these approaches, binarization in different color channels was also not explored. Chen et al. [20] proposed video text recognition using error voting and sequential Monte Carlo. This method depends on segmentation of character and thresholds. Text recognition from video frames, proposed in [41], mainly focused on character contours restoration. This approach used positive and negative peaks of Laplacian operators to detect the text in video frames and then used symmetry properties of edge pixels for identification of probable spatial candidate pairs followed by recognition algorithm for restoring the complete contours of character components. The work in [40] introduced a generalized approach based on fraction calculus to reduce the noise in images introduced by Laplacian operators. Roy et al. [27] proposed an approach to recognize the video text by Bayesian classifier through binarizing the image. To the best of our knowledge, existing methods have not yet explored the color channel specific texture properties for word recognition in the video frames and scene images. In this paper, we avoid the traditional complex binarization method by using a feature extraction procedure based on proper color channel selection.

## 3. Proposed Framework

In our work, we considered segmented word from scene text image or video text frame as an input to our framework. There exist a number of works [3, 12] for text localization and segmentation. In this framework we considered the approach [3] to get the segmented word images for our experiment. Synthetic fonts are used to train the algorithm along with maximum stable regions to segment the word images that shows robustness to geometric and illumination condition. Here, we consider only the word images to study our framework as most of the existing methods have evaluated their performance on word images.

We introduce a novel feature extraction method for color scene/video images in this paper. Most of the existing work on retrieval of words from scene/video images mainly concentrates on binarizing the images in the first step and then features are extracted from those binarized images. However, binarization is difficult in scene/video images due to a very uncertain variation of light and illumination of the images. Due to improper binarization, important information may get lost unintentionally which is a big hindrance for word retrieval in scene/video images. Here we have handled this constraint of binarization for video/scene images using a novel feature extraction approach. After this, we have put forward this color channel specific feature extraction method to sliding window level of feature extraction for word recognition where we found a significant improvement over the previous method of selecting a single color channel for a whole image. A flowchart of our feature extraction approach is given in Fig.2. We have found experimentally that a particular color image gets represented in a better way in one or more color channel than others. That means, those particular color channel(s) contains the better word texture information than the others.

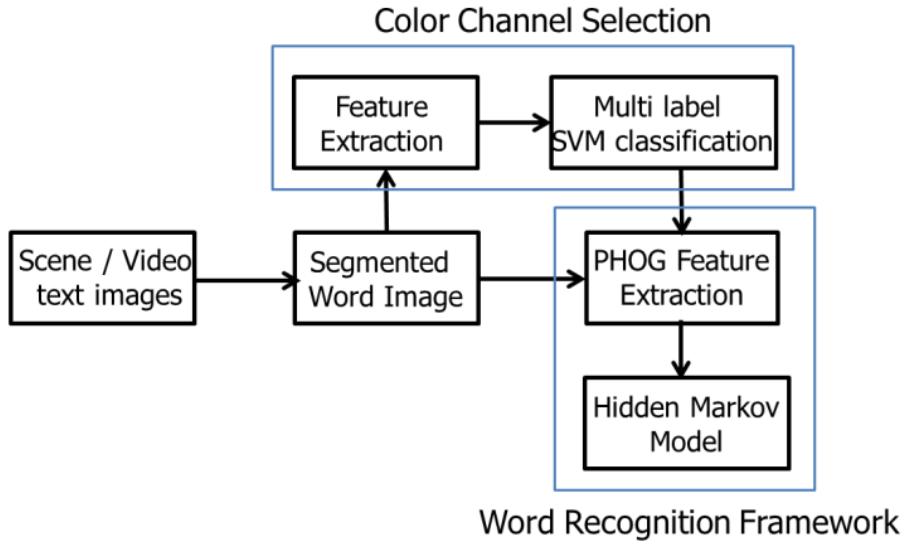

**Fig.2. Flowchart of our proposed framework**

## 3.2. Feature for Color Channel Selection

In machine learning approaches, a large amount of annotated dataset is required for training. For an automatic selection of color channel using multi-label SVM, the challenge is to get proper color channel information so that machine can learn which color channel better describes the properties in the images. We have made the labels for images through an automatic measure. From every dataset, we have considered 4- folds cross validation (see section 5.1 for more details). Then we perform the recognition task 8 times individually extracting the features from 8 color channels (R, G, B, Y, Cr, Cb, S, V). If correct result corresponding to ground truth is obtained for any image using a particular color channel, then we label +1 for that channel corresponding to the image and it is labelled with -1 if correct result is not obtained. Note that, we exclude the H channel due to very low recognition result obtained from this channel only (detailed in Section 5.3). The total correct results found using H channel can also be obtained using rest of the channels; same is not true for any other channels. Thus, we get an 8-bit string with either +1 or -1 value. We repeat the same procedure in a cross-validation way so that annotations of all images are obtained. The process of automatic preparation of transcription for color channel selection is explained using a diagram in Fig.3. The information due to H channel is provided in the diagram to provide a general overview of our color channel selection method.

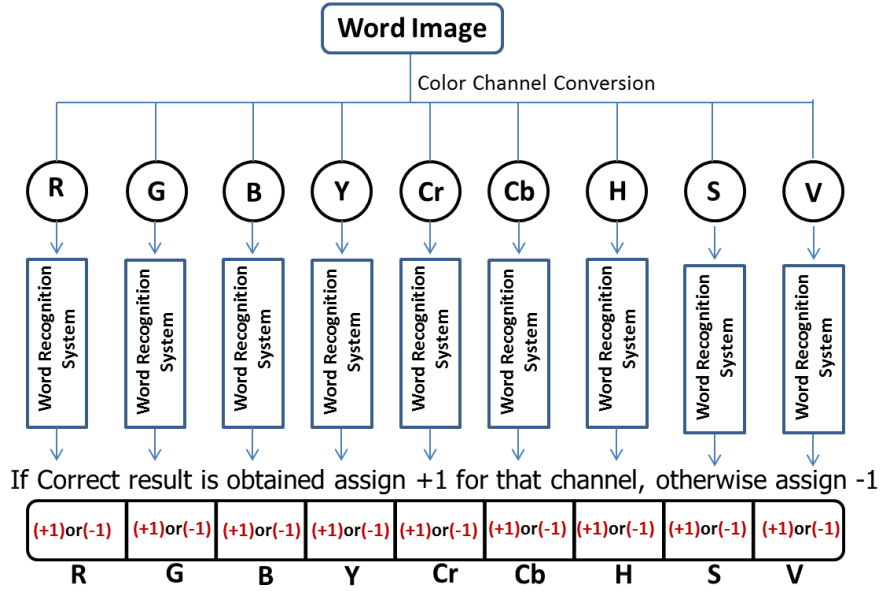

**Fig.3. Automatic preparation of transcription for color channel selection**

To extract the texture information from the color channel we have used image property based features here. Inspired by the idea of binarization algorithm selection [17], we have considered the feature used by the authors of [17] as a baseline. Though, there exist some texture based descriptors [22-24, 54, 56-59], in this paper we have used four different texture features namely wavelet, Gabor, LBP and LPQ for color channel selection along with the feature described in [17[. Descriptions of these features are given below. We extracted those features from individual color channels and the next the features are concatenated to form the final feature vector with normalization.

**3.2.1. Wavelet Feature:** In this paper, we used discrete wavelet transform (DWT) [24] feature for the selection of appropriate color channel along with other features. The statistical Mean and the Standard Deviation are used for feature extraction from each transformed images. The images are divided into 4 sub-bands i.e. LL, LH, HL, HH. LL is the approximation image i.e. coarse level coefficient while LH, HL and HH are the detail image i.e. finest wavelet coefficient. Only LL is critically subsampled by applying DWT again and again until the final scale is reached. The example of the two-level decomposition of an image is given in Fig.4.

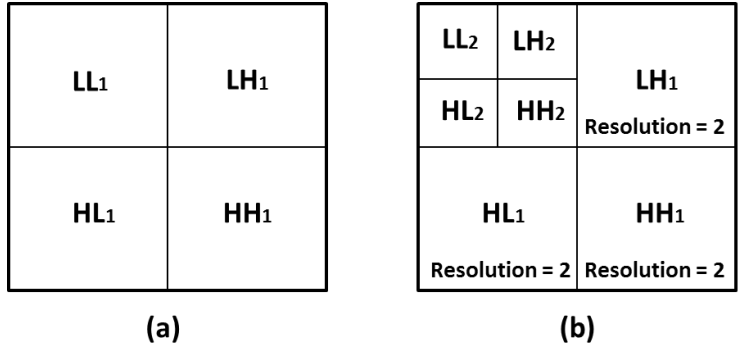

**Fig.4. Image decomposition. (a) First level (b) Second level.**

The transformed coefficient or value obtained is essential features which help in the selection of a color channel. So, the sub-bands of three-level decomposed images (i.e., $LL_k$, $LH_k$, $HL_k$, and $HH_k$; for k = 1, 2, 3, 4) are calculated as features using the formulas given in the equations (1) and (2) respectively.

$$mean\ (m) = \frac{1}{N^2} \sum_{i,j=1}^{N} p(i,j) \ldots\ldots\ldots (1)$$

$$Standard\ deviation\ (S_d) = \left[\frac{1}{N^2} \sum_{i,j=1}^{N} [p(i,j) - m]^2\right]^{\frac{1}{2}} \ldots\ldots\ldots (2)$$

Where $p(i,j)$ is the transformed value in $(i,j)$ for any sub band of size $N \times N$.

**3.2.2. Gabor Filter:** Local power spectrum based features [22] of images has been used for texture analysis, which is established through a feature that is obtained by using 2-dimensional global filters which filter the input image. These filters are local and linear. For our work, we have used Gabor energy and Gabor mean as features for the selection of a color channel. A bank of Gabor filters is used to extract the local image features. An input image $I(x, y), (x, y) \in \Omega$ (where $\Omega$ is the set of image points) is convoluted with a 2-D Gabor function $g(x, y), (x, y) \in \Omega$ to obtain a Gabor feature image $r(x, y)$ as follows:-

$$r(x,y) = \iint_{\Omega} I(\xi, \eta) g(x - \xi, y - \eta) d\xi d\eta \ldots\ldots\ldots (3)$$

For our work, we have used low pass filter for filtering the images along with 5 scales and 6 orientations.

**3.2.3. Local Binary Pattern (LBP):** Local binary pattern images [23] have been used for image classification of gray level as well as color space. Let, $V_{x,y}$ is the value of the pixel which is located at the position$(x, y)$ in an image where $x \in 1,2,3,\ldots H$ and $y \in 1,2,3,\ldots,W$. Here, H and W are height and width of the images in spatial domains. For color image, the value of a pixel is represented by vector $s = (C_1, C_2, C_3)$, where $C_i$ is the color channel of a color space. This $V_{x,y}$ value can be used to obtain the total order of vector by using the partial order method of porebski such that $S_1$ proceeds $S_2$ if $S_1 < S_2$. LBP is an operator for image description that is based on the signs of differences of neighboring pixels [23].

**3.2.4. Local Phase Quantization (LPQ):** The LPQ [54] operator is used for color channel selection by computing LPQ locally at every pixel location and resulting code in form of histogram. Suppose $f(x)$ is the original image, $g(x)$ is the observed image and $h(x)$ is the point spread function. The discrete model of spatial invariant blurring of observed image is defined by convolution as

$$g(x) = f(x) \otimes h(x) \quad \ldots \ldots \ldots (4)$$

Where, $\otimes$ denotes the 2-D convolution and x is the vector of position $[x, y]T$.

**3.2.5. Feature for Binarization Selection:** Chattopadhyay et al. [17] proposed an image specific binarization algorithm selection using image properties like mean, variance and skewness and histogram feature from Hue channel. Here, we have used similar image features from color channels. From RGB color space, three features mean, variance and skewness are calculated using following formulas for each color channels R, G, and B.

$$E_x = \frac{1}{N} \sum_{y=1}^{N} S_{xy} \quad \ldots \ldots \ldots (5)$$

$$\sigma_x = \left[ \frac{1}{N} \sum_{y=1}^{N} (I_{xy} - E_x)^2 \right]^{\frac{1}{2}} \quad \ldots \ldots \ldots (6)$$

$$z_x = \left[\frac{1}{N}\sum_{y=1}^{N}(I_{xy} - E_x)^3\right]^{\frac{1}{3}} \quad \ldots\ldots\ldots (7)$$

where, $E_x$ is the mean, $\sigma_x$ is the variance, and $Z_x$ is the skewness of each color plane having segment $S_{xy}$. $I_{xy}$ is the value of x color space at y pixel and N is the total number of pixels. A color histogram with 256 bins is also considered that is invariant to the transformation [17]. For this purpose, we used HSV (16, 4, 4) quantization scheme for histogram feature. Finally, a feature vector of length 265 (256 + 3×3) is extracted from each video /scene word image.

### 3.3. Color Channel Selection using SVM

Multi-label classification problem [2] can be handled by transforming the problem into a set of independent binary classification problems using "*one vs all*" scheme. Given a multi-label training dataset D = {($x_1$, $y_1$), ($x_2$, $y_2$),…,($x_n$, $y_n$)}; where $x_i$ is the input feature vector and $y_i$ is the label vector of input $x_i$ having value either +1 or -1. Each vector can be of length K which is equal to the number of classes. In our system, K is considered as 8 because 8 channels were considered. $Y_i$ is the label vector for $i^{th}$ word image for color channel selection which is obtained through an automatic measure as mentioned in Section 3.2 (see Fig. 3). $Y_{ik}$ = +1 indicates that $i^{th}$ data is assigned to $k^{th}$ class. The standard quadratic optimization problem for SVM training is defined as:

$$\min_{W_k, b_k, \{\xi_{ik}\}} \frac{1}{2}||W_k||^2 + C\sum_{i=1}^{N}\{\xi_{ik}\} \quad \ldots\ldots\ldots (8)$$

This is subjected to the constraints $y_{ik}(W_k^T x_i + b_k) \geq 1 - \xi_{ik}$, $\xi_{ik} \geq 0, \forall i$; where $\{\xi_{ik}\}$ are the slack variables, C is the trade-off parameter which maximizes the soft class separation margin $W_k$ and $b_k$ are model parameters which define a binary classifier associated with k-th class. The binary classifier associated with k-th class: $f_k(x_i) = W_k^T x_i + b_k$. The prediction of the label vector $\hat{y}$ for an unlabelled instance $\hat{x}$ is performed using the set of binary classifiers from all classes. The k-th component of the predicted label vector is +1, if $f_k(\hat{x}) > 0$; otherwise it will be -1. The absolute value of predictive function $|f_k(\hat{x})|$ provides the confidence value for its prediction $\widehat{y_k}$ corresponding to the given input test feature vector. During text recognition in HMM framework, the color channel with highest confidence value is considered as the final selection. If more than one channel is having same (maximum) confidence value, any one of them is selected.

## 4. Word Recognition

We have used HMM-based framework for word recognition in scene text images and video frames. Sliding window based PHOG features [19] are extracted from the word image and these features are then fed into HMM for text recognition. The reason behind considering each individual sliding window for color channel selection than considering a whole word image is that there exist varying illuminations over different regions of a single word image. A single color channel may not be a good option for all the sliding window patches of an image. So, a color channel is selected for each individual sliding window and PHOG feature is extracted using that particular color channel. A similar patch based feature extraction method has been studied in [45]. Sliding window based word recognition framework is shown diagrammatically in Fig.5. Details of word recognition framework are mentioned in following subsections.

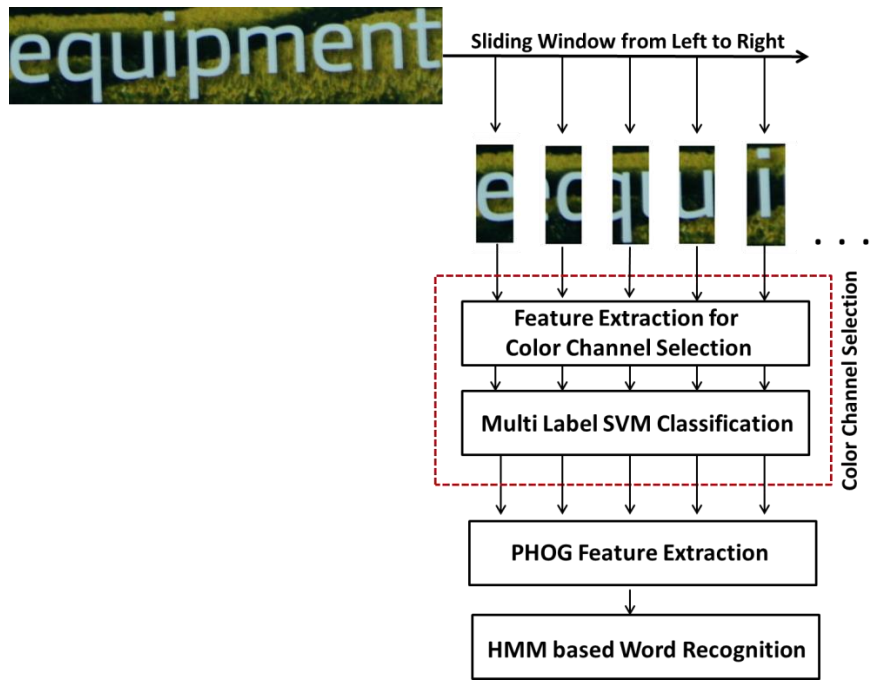

**Fig.5. Sliding window based word recognition framework**

## 4.1. Feature Extraction for Word Recognition

For feature extraction, a sliding window of width 40×8 is being shifted from left to right with 50% overlapping between two sliding windows. We have normalized the height of each word image to 40 keeping the aspect ratio same. To take care of multi-oriented text, this movement of sliding window follows the polynomial equation fitted for the multi-oriented word image as mentioned in [1]. A general polynomial equation is given by

$$f(x) = a_n x^n + a_{n-1} x^{n-1} + \ldots\ldots\ldots + a_1 x^1 + a_0 \ldots\ldots\ldots (9)$$

Where $n > 0$ and $a_0, a_1, \ldots a_n$ are the positive real number which is evaluated by the curve fitting algorithm. We use second-degree polynomial using $n = 2$. The height of the word image is found taking the average of the height from each point of the polynomial equation.

**Sliding Window Feature:** PHOG [47] is the spatial shape descriptor which gives the feature of the image by spatial layout and local shape comprising of gradient orientation at each pyramid resolution level. PHOG features have been used for sliding window based feature extraction for HMM-based handwritten word recognition in [19]. To extract the feature from each sliding window, we have divided it into cells at several pyramid level. The grid has $4^N$ individual cells at $N$ resolution level (i.e. $N = 0, 1, 2..$). Histogram of a gradient orientation of each pixel is calculated from these individual cells and is quantized into $L$ bins. The concatenation of all feature vectors at each pyramid resolution level provides the final PHOG descriptor. At any individual level, it has $Lx4^N$ dimensional feature vector where N is the pyramid resolution level (i.e. $N = 0, 1, 2..$). In our implementation, we have limited the level (N) to 2 and we considered 8 bins (360º/45º) of angular information. Increasing the value of N beyond 2 does not increase the accuracy much but it increases the computation time of word recognition framework due to large feature dimension. So we obtained $(1 \times 8) + (4 \times 8) + (16 \times 8) = (8+32+128) = 168$ dimensional feature vector for individual sliding window position.

## 4.2. Word Recognition using Hidden Markov Model

The text recognition is performed based on processing the feature vector using left to right continuous density HMM [1, 19]. The major reason behind choosing HMM is that it can model sequential dependencies. An HMM can be defined by initial state probabilities $\pi$, state transition matrix $P = [p_{ij}]$, $i, j=1,2,...,N$, where $p_{ij}$ denotes the transition probability from state $i$ to state $j$ and output probability $F_j(O_k)$ modeled with continuous output probability density function. The density function is written as $F_j(x)$, where x represents *k* dimensional feature vector. A separate Gaussian mixture model (GMM) is defined for each state of model. Formally, the output probability density of state *j* is defined as

$$F_j(x) = \sum_{k=1}^{M_j} c_{jk}\, \mathcal{N}(x, \mu_{jk}, \Sigma_{jk}) \quad \ldots\ldots\ldots (10)$$

where, $M_j$ is the number of Gaussians assigned to *j*. and $\mathcal{N}(x, \mu, \Sigma)$ denotes a Gaussian with mean $\mu$ and covariance matrix $\Sigma$ and $c_{jk}$ is the weight coefficient of the Gaussian component k of state *j*. For a model $\lambda$, if $O$ is an observation sequence $O = (O_1, O_2,..., O_T)$ which is assumed to have been generated by a state sequence *Q= (Q₁, Q₂,..,Q_T)*, of length *T*, we calculate the observation's probability or likelihood as follows:

$$P(O, Q|\lambda) = \sum_Q \pi_{q1} F_{q1}(O_1) \prod_T p_{qT-1\ qT}\, F_{qT}(O_T) \quad \ldots\ldots\ldots (11)$$

Where $\pi_{q1}$ is the initial probability of state 1.

The feature vector sequences along with the transcriptions of the text line are used to train the character model of HMM. Text line model is produced after concatenating the character models. The recognition is performed using Viterbi algorithm which finds the best likely hood character sequence for a given feature vector sequence. We used HTK toolkit for our implementation.

## 5. Experiment and Results

In this section, we report the performance of our color channel selection based word recognition framework. We first introduce the datasets and then present the detailed recognition result along with different error analysis and discussions.

### 5.1. Dataset

To evaluate the performance of our color channel selection based word recognition framework, we have performed the experiment for both scene text and video frames. Our word recognition system consists of two steps. At first, a color channel is selected using SVM classifier and next, word recognition is performed in that color channel. We have considered 4 fold cross validation for color channel section process. In 4-fold cross validation, 1 fold (25% data) is kept for test and remaining 3 folds (75% data) are used for training. The process is repeated 4 times (considering four different 1-folds as testing and corresponding remaining 3-folds as training) and finally an average of these results was reported in the paper. For publicly available datasets, we use the division provided in the corresponding dataset for training and testing purposes. In case of our Indic dataset (i.e. text recognition for Devanagari word images), we use 5-fold cross validation technique, i.e. 4 folds are used as training and rest 1-fold is used as testing. To optimize different parameters (e.g. cost parameter of multi-label SVM, State number and Gaussian number of HMM) of the experiment, we have prepared a validation dataset. This validation dataset comprised of some examples from different datasets considered in our experimental study.

To check the generality of our feature extraction approach we considered both English and Non-English datasets. ICDAR 2003[13] is one of the most popular scene text image recognition dataset for Latin (English) script. ICDAR 2013 video text dataset is considered for word recognition in video frames which contains a total of 28 videos. We collected a total of 589 scene images and 40 videos from news-channels collected from YouTube for Devanagari script. Out of these, 4,947 word images from 589 images have been considered for scene text image recognition and 4,848 word images are considered for video text word image recognition, respectively. Few examples of our segmented word images and scene/video images are given in Fig.6 and Fig.7 respectively. Please note that segmented word images are considered as the input to our word recognition system. We have considered the lexicon size of 10K for each of the experiments where words are collected from available online newspapers.

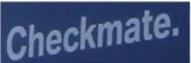

**Fig.6. Examples of word English and Devanagari images considered in our datasets (both horizontal and multi-oriented)**

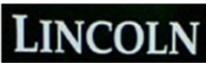

**Fig.7: Examples of video frames and scene images considered in our experiment.**

## 5.2. Color Channel Selection Performance

To evaluate the performance of our multi-label color channel selection framework, we have used the evaluation matrices [46]. In case of multi-label classification, a single sample may belong to one or more classes. Thus, we used different metrics than those used in traditional single-label classification. Here, let N be the total number of sample data in a dataset, $Y_i$ be the ground truth label of $i^{th}$ data sample which is a $K$ bit binary string containing each bit as either +1(positive instance for that class) or -1 (negative instance for that class). $Z_i$ is the predicted multi-label SVM output having same size as $Y_i$. We have prepared the annotation for all the word images by automatic measure as explained in Fig. 3(in Section 3.2). The following measures [46] are used to evaluate the performance of different features for our SVM based multi-label classification problem.

$$Accuracy = \frac{1}{|N|} \sum_{i=1}^{N} \frac{Y_i \cap Z_i}{Y_i \cup Z_i} \ldots\ldots\ldots (12)$$

$$Precision = \frac{1}{|N|} \sum_{i=1}^{N} \frac{Y_i \cap Z_i}{Z_i} \ldots\ldots\ldots (13)$$

$$Recall = \frac{1}{|N|} \sum_{i=1}^{N} \frac{Y_i \cap Z_i}{Y_i} \ldots\ldots\ldots (14)$$

We have evaluated these measures for ICDAR 2003 dataset and Devanagari scene image dataset with respect to different features used in our color channel selection purpose. The evaluation matrices of multi-label classification results for color channel selection are given in Table I. Wavelet feature provided best results among all other features. We have tested our color channel selection framework at different values of cost parameter (C). These values include the range from 0.1 to 1 with an interval of 0.1, from 1 to 10 with an interval of 1 which is followed by the range 10 to 100 with an interval of 10. The optimum value of cost parameter (C) in multi-label SVM classification has been found experimentally as 1.

Table I: Performance of multi-label classification for color channel selection

| Feature for Color Channel Selection | English(ICDAR 2003) | | | Devanagari | | |
|---|---|---|---|---|---|---|
| | Accuracy | Precision | Recall | Accuracy | Precision | Recall |
| Wavelet Feature | **79.42** | 85.36 | 83.29 | **78.94** | 84.69 | 84.69 |
| Gabor Feature | 75.21 | 83.69 | 83.14 | 75.02 | 82.36 | 82.39 |
| Chattopadhyay et al.[17] | 73.36 | 81.36 | 80.01 | 72.61 | 80.42 | 81.03 |
| LBP Feature | 70.58 | 78.36 | 78.16 | 71.18 | 79.08 | 77.64 |
| LPQ Feature | 69.48 | 75.69 | 75.81 | 68.36 | 74.26 | 73.98 |

## 5.3. Evaluation of Text Recognition

For HMM-based word recognition, we have evaluated the performance of our framework at different state numbers from 4 to 8 with a gap of 1. For Gaussian number, it has been varied from 8 to 128 with a step of power of 2. Here, we have noted that state number 6 and Gaussian number 32 provide best results in the validation dataset. Hence we set these parameters for word recognition framework. We have evaluated the word recognition performance in all the 9 color channels using PHOG feature extracted from an individual color channel. Please note that, we have considered the color channel having highest confidence value. We did a preliminary experiment with H channel of HSV. From the experiment we obtained only 8.87% accuracy in ICDAR 2003 dataset. We also noted that the correct results obtained using H channel can be also obtained by other channels. If H channel was considered, the runtime performance would be increased without any improvement in recognition performance. Hence, this H channel was not considered. However, this is not the case for Cr and Cb channel. Though, some results obtained through these channels are poor, but in some scenarios, these channels provide correct recognition results which other channels fail to do so. Hence, the final recognition performance gets improved by including these Cr and Cb channels. To prove the complementary nature among the color channels an Oracle approach is performed over all the results obtained from each color channel for a word image. The Oracle approach indicates the maximum limit up to which the recognition accuracy can be achieved using proper color channel selection approach. Suppose, a word image is recognized correctly using a particular color channel but it fails to give correct result with other color channels. With oracle approach, the result will be treated correct, since, correct result has been found by any one of these color channels. However, with color properties analysis, the proper color channel might not be found always. Some qualitative results in ICDAR 2003 datasets are shown in Fig.8. Here, traditional approach stands for global histogram based Otsu binarization method followed by word recognition using OCR [18] as used in [1]. Due to poor binarization results, the traditional way of recognition does not provide correct results whereas, our color channel selection based word recognition framework recognizes those word images correctly.

| Word Images | Binarized Image | Traditional Approach | Proposed Approach | Image at Selected Color Channel |
|---|---|---|---|---|
| limited | limited | ✗ | ✓ (R) | limited |
| AMERICAN | AMERICAN | ✗ | ✓ (Y) | AMERICAN |
| Start | | ✗ | ✓ (Cb) | Start |
| STATE | | ✗ | ✓ (Cr) | STATE |
| twitter | twitter | ✗ | ✓ (G) | twitter |
| Coke | Coke | ✗ | ✓ (V) | Coke |
| Masterfile | Masterfile | ✗ | ✓ (G) | Masterfile |
| Jewelers | Jewelers | ✗ | ✓ (S) | Jewelers |

**Fig.8. Qualitative study of word recognition performance**

The quantitative recognition result using different individual color channel is given in Table II. From the experiment results, we found a significantly improved recognition performance than using a single color channel. This ensures that few color channels describe the word texture information better than others. Our proposed color channel selection approach can uplift the recognition performance to 78.44% in ICDAR 2003 scene image dataset. The recognition result in the video frame is 75.41% in ICDAR 2013 video text dataset which is due to better resolution of the videos compared to scene image dataset. To check the script independency of the dataset we checked our recognition framework in Devanagari script for both scene image and video frame. We got 72.87% accuracy in scene image dataset whereas 71.14% for video frames. This comparatively low result is due to the presence of many characters and modifiers in Devanagari script [19].

Table II: Word recognition performances

| Color Channel | English | | Devanagari | |
| --- | --- | --- | --- | --- |
| | Scene Image (ICDAR 2003) | Video Frame (ICDAR 2015) | Scene Image | Video Frame |
| R | 63.88 | 62.87 | 58.94 | 59.18 |
| G | 68.87 | 67.48 | 63.24 | 62.96 |
| B | 66.48 | 65.69 | 61.28 | 60.14 |
| Y | 66.14 | 65.14 | 61.48 | 62.86 |
| Cb | 28.98 | 27.89 | 23.58 | 24.29 |
| Cr | 31.47 | 32.89 | 24.89 | 26.97 |
| H | 8.87 | 9.89 | 7.15 | 6.48 |
| S | 48.98 | 47.96 | 42.18 | 44.69 |
| V | 65.89 | 65.97 | 60.89 | 61.48 |
| **Oracle Approach** | **82.78** | **79.28** | **76.48** | **75.36** |
| **Proposed Approach** | **78.44** | **75.41** | **72.87** | **71.14** |

A comparative performance of recognition performance in using different channel has been done graphically for ICDAR 2003 scene image dataset in Fig.9 where we have excluded the H channel result because of its very low performance. In Fig.10, a comparative performance is shown in our proposed approach, binarization (using wavelet gradient) based approach [1] and Oracle approach using PHOG feature for word recognition. The qualitative results obtained using different color channels are shown in Fig.11.

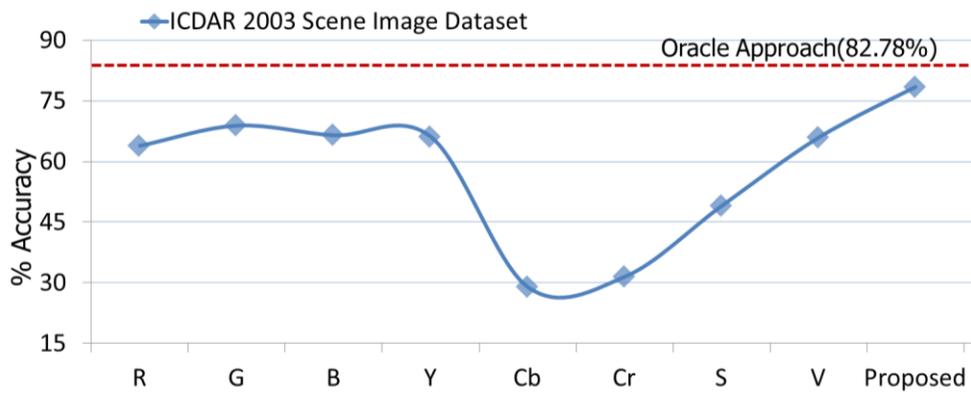

**Fig.9. Comparative study of word recognition performance using different color channels for feature extraction.**

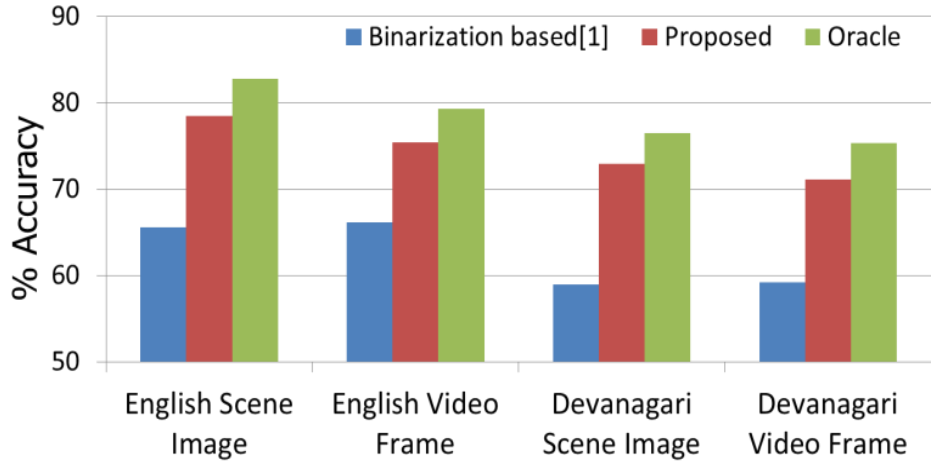

**Fig.10. Comparative study showing the improvement over traditional approach and the result obtained using Oracle approach.**

| Word Image | Binarized Image | Binary | R | G | B | Y | Cr | Cb | S | V | Proposed |
|---|---|---|---|---|---|---|---|---|---|---|---|
| LION | LION | WON | YOU | IION | LION | HEN | IN | INK | LIE | LIO | LION |
| Times | Times | His | Times | Ties | Timex | Timex | Tint | TI | Titan | Home | Times |
| Tree | Tr | The | here | hen | Tree | The | Bee | Tea | True | Toll | Tree |
| BANK | BANK | BAR | READ | BAND | BANK | Burst | RISK | By | BAND | BANK | BANK |
| LOAD | LOAD | Boat | HOLD | HOLD | LEAD | LOAD | LOOK | HOLY | LOAD | Loaf | LOAD |
| ROAD | ROAD | Loot | READ | ROAD | ROAD | ROAL | READ | RIDE | READ | BED | ROAD |
| BEST | BEST | RAT | BEST | LAST | REST | BELL | BALL | BEST | RESET | REST | BEST |
| WILL | WILL | kill | WILL | WILL | kill | vibe | WILL | BILL | BILL | kill | WILL |

**Fig.11: Qualitative result showing word recognition result obtained in different color space (Red marked text denotes wrong result and green marked text denotes correct result).**

Here, we have used color channel selection technique for individual sliding window to handle the different illumination condition over a single image. By considering color channel selection in each sliding window patches than whole image, the improvement has been found from 72.48% to 78.44% in ICDAR 2003 dataset. Few images are shown in Fig. 12 where color channel selection for a single whole image fail to give the correct recognition result, whereas sliding window based color channel selection method gives the correct recognition result.

To test the performance of our word recognition system at varying illumination condition, we have collected a total of 312 word images of such varying illuminations. These 312 word images are collected from different datasets and few are synthetically generated. We consider the model trained from ICDAR 2003 dataset for testing in case of these images. We noticed an accuracy of 75.32% while selecting color channel at sliding window level. The result is reduced to 64.74% if a single color channel is considered for the whole word image. Under the same experimental condition, the result using the method of [1] is only 53.84%.

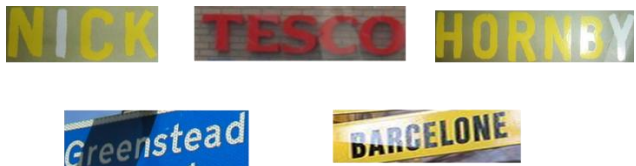

**Fig.12: Word images with varying illumination where our sliding window based color channel selection method gives correct results but if we had considered a single color channel for the whole image, it fails.**

We have evaluated our framework for different resolutions of the word images. For this purpose, we have considered reducing the resolution of the images by reducing the height of the word images from its original height to its 20% value with a gap of 10% by keeping the aspect ratio same. The recognition accuracy for ICDAR 2003 dataset at different resolution levels is shown in Fig13(a). We noted that the recognition performance drops from 60% reduction. Some sample images at different resolutions are shown in Fig.13(b).

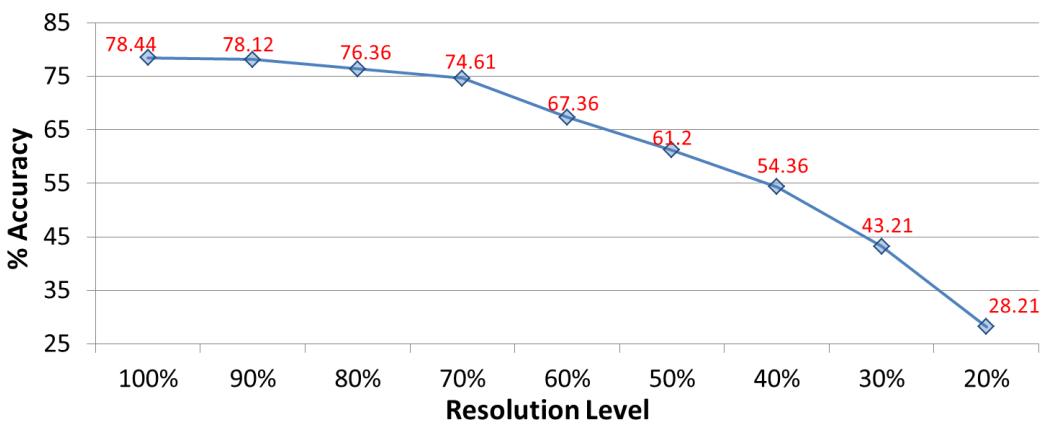

(a)

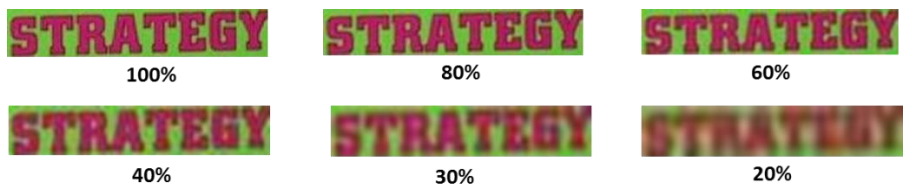

(b)

**Fig.13: (a) Recognition accuracy at different resolution level in ICDAR 2003 dataset. (b) Qualitative example of images at different resolution levels are shown.**

## 5.4. Comparative Study of Features

In this paper, we have considered five different features for evaluation of performance for color channel selection. We considered Wavelet feature, Gabor feature, LPQ feature, LBP feature along with the feature used by Chattopadhyay et al. in [17] for binarization algorithm selection. Among these features, wavelet features outperforms others in terms of recognition accuracy. The recognition performance using different features is given in Table III. A comparative study between PHOG and LGH (Local Gradient Histogram) features [23] for word recognition is also studied in Table IV and Table V where we considered the few public datasets for both video and scene images, respectively. For video datasets, we considered ICDAR 2013, ICDAR 2015 and YouTube Video Text (YVT) dataset. The datasets considered for scene image are ICDAR 2003, MSRA-TD500, SVT, ICDAR 2013, IIIT5K words. It was noted that PHOG features outperforms the LGH features in recognition performance.

Table III: Word recognition performances using different features for color channel selection

| Feature for Color Channel Selection | English | | Devanagari | |
|---|---|---|---|---|
| | Scene Image | Video Frame | Scene Image | Video Frame |
| Wavelet Feature | 78.44 | 75.41 | 72.87 | 71.14 |
| Gabor Feature | 76.51 | 73.97 | 70.14 | 69.01 |
| Chattopadhyay et al.[17] | 74.14 | 71.94 | 67.17 | 66.14 |
| LBP Feature | 72.01 | 69.89 | 66.47 | 65.98 |
| LPQ Feature | 71.18 | 68.47 | 65.78 | 65.11 |

Table IV: Comparative study between LGH and PHOG feature in standard video datasets

| Datasets | PHOG | | LGH | |
|---|---|---|---|---|
| | Character | Word | Character | Word |
| ICDAR-2013[33] | 85.47 | 75.41 | 83.69 | 73.22 |
| ICDAR 2015[37] | 88.68 | 79.48 | 86.47 | 77.41 |
| YVT[39] | 89.12 | 81.24 | 86.71 | 78.94 |

**Table V: Comparative study between LGH and PHOG feature in standard scene datasets**

| Datasets | PHOG | | LGH | |
|---|---|---|---|---|
| | Character | Word | Character | Word |
| ICDAR-2003[29] | 86.47 | 78.44 | 84.11 | 75.94 |
| MSRA-TD500[34] | 85.69 | 76.14 | 83.01 | 73.69 |
| SVT[30] | 87.94 | 77.24 | 85.11 | 75.58 |
| ICDAR 2013[33] | 90.87 | 82.31 | 89.24 | 80.12 |
| IIIT5K[32] | 89.12 | 80.12 | 87.69 | 78.61 |

## 5.5. Comparison with Existing Approaches

Word recognition performance using other popular binarization algorithms is also studied in [1]. In Fig. 14, we compare recognition performance considering different binarization methods followed by an available OCR system [18] and our HMM-based recognition framework. There exist many works [15, 32, 35, 36] towards word recognition in scene and video frames. In Table VI, a comparative study with some existing approaches is given. ABBY FineReader [4] converted images into editable file format with poor accuracy. Wang et al. [15] used multi-scale sliding window classification along with non-maximum character suppression to obtain the character sequences. Mishra et al. [32] presented a framework which used both bottom-up (character) and top-down (language) cues for text recognition. They considered Conditional Random Field to jointly model strength of sliding window based text detection and association among them. Simultaneous modeling of both visual and lexicon consistency of words in a single probabilistic model was explored in [35]. Khare et al. [11] proposed an end to end system using blind deconvolution model to enhance the edge intensity of the blurred pixel for text detection and recognition. A learned representation called Strokelets was used to capture the essential substructures of characters in [36]. Our system outperforms most of the other methods with same lexicon size. In ICDAR2003 dataset with lexicon-50, the recognition result due to Yao et al.[36] was little better than us. The recognition performance obtained in [36] in IIIT5K dataset are 80.2% and 69.3% using lexicon size of 50 and 1K respectively. We performed the experiment with a larger lexicon size of 10K and achieved a recognition accuracy of 80.12%. The lexicon sizes of datasets are marked in Table VI. Some of the methods have not provided results in some of these datasets, so those places in Table VI are left blank. Recently, deep learning based approaches [49, 50, 51] have been applied on scene text recognition and achieved better performance. The superior performance of these methods heavily depends on large amount of training data. For example, the method in [50] was trained on about 14k words and 71k characters from Flickr. About 7.2 million synthetically generated data was used in [49]. However, our proposed method did not use any synthetic data and only used the training example provided in each dataset. Moreover, conventional approaches may provide some interesting properties which can be complementary to deep learning based method.

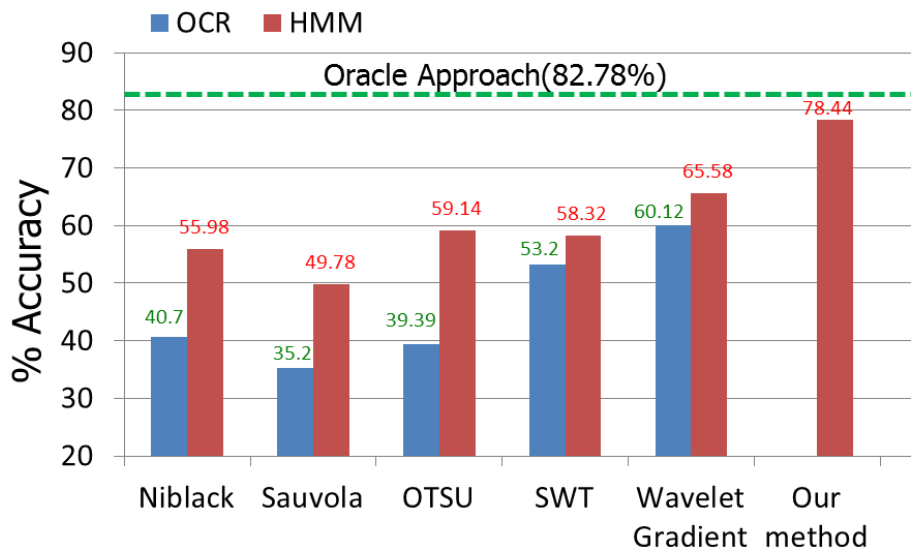

**Fig.14: Comparative study of HMM and OCR based recognition result in ICDAR 2003 dataset**

Table VI: Comparative study of different approaches on standard scene text recognition dataset

| Datasets | ICDAR 2003 (Lexicon-50) | ICDAR 2003 (Lexicon -1156) | SVT (Lexicon 50) | ICDAR 2013 | IIIT-5K (Lexicon-50) | IIIT-5K (Lexicon-1K) |
|---|---|---|---|---|---|---|
| Baseline ABBYY[4] | 56.0 | 55.0 | 57.0 | - | 35.0 | 24.3 |
| Wang et al. [15] | 76.0 | 62.8 | 57.0 | - | - | - |
| Mishra et al. [32] | 81.8 | 67.8 | 73.2 | - | 64.1 | 57.5 |
| Novikova et al. [35] | 82.8 | 72.9 | - | - | 64.1 | 57.5 |
| Yao et al. [36] | 88.0 | 80.3 | 75.9 | - | 80.2 | 69.3 |
| Khare et al. [11] | - | - | - | 74.2 | - | - |
| Proposed | 86.92 | 82.97 | 82.36 | - | 86.21 | 84.37 |
| Proposed (10K lexicon) | 78.44 | | 77.24 | 82.31 | 80.12 | |

## 5.6. Error Analysis and Discussions

Few word images are shown in Fig.15 where our proposed approach did not work properly. Fig. 15(a) shows a sample example where our system failed to give the correct result due to very low resolution and poor edge information. Our system may not work in case of too much stylistic fonts (See Fig. 15 (b)). Since such fonts were not trained properly through character-based HMM models, stylistic character recognition may not work. Excessive color variation is also a reason for wrong recognition result as shown for Fig 15(c). Sometimes, it may happen that the color channel selection framework fails to give the best color channel for feature extraction in sliding window. In such scenarios, our framework may also fail to provide the correct recognition result.

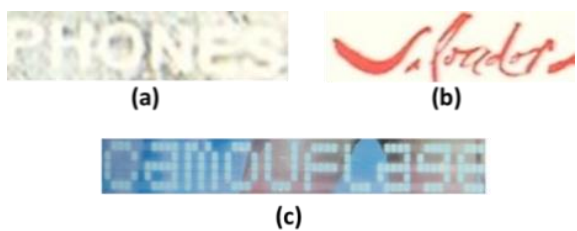

**Fig.15: Some examples are shown where our framework failed to provide correct recognition result.**

## 5.7. Experiment with Noisy Images

We have tested our word recognition framework with the word images added with synthetic noises. We have added Gaussian noise of different levels to see the effect of noise on our framework. We have used noise levels 5% to 30% with a successive interval of 5%. Some qualitative results are shown with noise (See Fig. 16(b)) of different levels with respect to the original word image shown in Fig. 16(a). Recognition accuracy falls from 78.44% to 72.01% when we add 30% Gaussian noise to the original word images in ICDAR 2003 datasets. Under similar experimental condition, the performance of binarization based method [1] is evaluated. Recognition accuracy of both our proposed method and binarization based method [1] with respect to different noise levels are shown in Fig.16(c). Along with Gaussian noise, we have also considered 'Salt and Pepper' and 'Speckle' noises by varying the noise levels from 0% to 30% with an interval of 5%. We found a drop of 7.3% and 6.67% at 30% noise level with respect to zero level for 'Salt and Pepper' and 'Speckle' noises respectively.

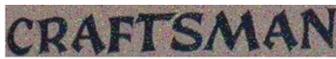

(a)

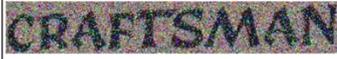  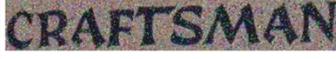

(i)                                (ii)

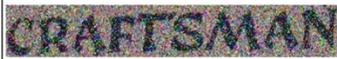

(iii)

(b)

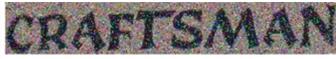

(c)

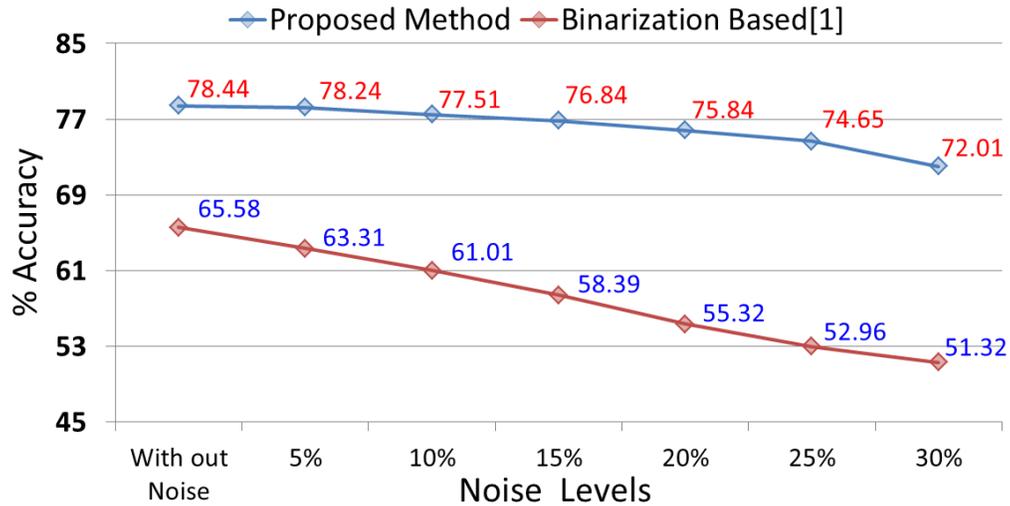

(d)

**Fig.16: Example showing word images degraded with different types of noise in (a) namely, Gaussian, Salt and Pepper and Speckle, respectively, with respect to the original word images shown in (a). Word images degraded with different levels of Gaussian noise are shown in (c). (d) shows the recognition accuracy in ICDAR 2003 dataset with different levels of added Gaussian noise by the proposed method and binarization method.**

**5.8. Runtime evaluation**

The framework has been studied using a computer I5 CPU of 2.80 GHz and 4GB RAM 64 bit. The average runtime has been computed from different runs made in the experiment. The average time for word recognition without color channel selection is 0.92 second. Out of this 0.92 second, 0.48 second was consumed for PHOG feature extraction and rest of 0.44 second is to get the recognition result from trained HMM model. Whereas using color channel selection process, the average time for word recognition has been increased to 1.61 seconds. The additional times was taken due to color channel selection based technique.

# 6. Conclusion

In this paper, we explored the text recognition performance in scene images and video frames by avoiding complex binarization algorithms. We propose a color channel selection framework for word recognition in script-independent and multi-oriented text. In this work, we have shown how the color channel specific information can be used for text recognition in scene images and video frames. The performance of the framework has been evaluated in both Latin (English) and Devanagari scripts. The recognition results obtained in different datasets are encouraging. The experimental analysis shows that binarization step can be avoided by selecting a proper color channel of an image patch for feature extraction which contains more details about the text information.

In future, we plan to improve our color channel selection framework through joint learning framework to optimize the color channel selection iteratively. More color channels may also be explored to check the efficiency. The contrast, texture or structure information can also be combined with color information to improve the text detection. The efficiency of color channel selection can be taken forward for text localization in video or scene images. Besides, color channel selection based feature extraction can be used in different texture classification tasks. In our work, we considered a single color channel for feature extraction whereas this can be extended to multiple color channel selection for feature extraction and different weights can be assigned to the features of color channel depending on the importance of the color channel.